# When Should a Voice Agent Speak? End-of-Turn Detection in Turkish

Ahmet Tuğrul Bayrak, Fatma Nur Korkmaz, Mustafa Sertaç Türkel, and Nur Yağmur Tuncer

Data Science and Innovation, Ata Technology Platforms, Istanbul, Turkey
tugrul.bayrak@atptech.com

**Abstract.** A voice agent that waits for a fixed silence before replying either interrupts users who pause mid-utterance or responds late to users who have finished. This study examines the decision at the onset of silence, using turn ends and turn-internal pauses extracted from the Real-TurnTurk corpus of Turkish dyadic video calls. A fine-tuned Turkish TurnGPT model, a zero-shot Turkish language model and a deployed multilingual detector are compared, together with voice-activity timing and facial features, under recording-disjoint and speaker-disjoint evaluation with paired cluster bootstrap tests. On held-out recordings, the fine-tuned model separates turn ends from pauses better than both text baselines and is the best calibrated of the three. Timing features of the current speaker are similarly informative, and their combination with the fine-tuned text score gives the best timing-based results; with logistic regression, this combination also outperforms the same combination with either text baseline. In a latency simulation, adding the text score to a silence timeout shortens the mean response delay at the same interruption rate. On automatic Whisper transcripts, the fine-tuned model retains its advantage over the zero-shot model.

## 1 Introduction

Spoken interaction is organised in turns that alternate with gaps of only a few hundred milliseconds [1]. Many deployed voice agents still decide when to respond with a silence timeout: the user is assumed to have finished once no speech has been detected for a fixed time. A short timeout interrupts users who hesitate, and a long one delays every response [2, 3]. For this reason, some current agents query an end-of-turn model when silence begins and respond early only when the model indicates that the turn is complete [4].

Turn-taking models are typically scored at every word or every frame, although most of these positions lie within a turn. The decision that determines the behaviour of the agent is taken at a silence onset, where a finished turn has to be distinguished from a pause after which the speaker continues. At that point the agent can draw on the transcript up to the pause, the timing of the preceding conversation, the speaker's voice and, in video calls, the faces of both participants [5, 6].

Since Turkish is agglutinative and verb-final, the lexical evidence of completion is concentrated in the suffixes of a single word, and resources for Turkish spoken dialogue are limited. A text-based end-of-turn model for Turkish, TurnTurk-GPT, was introduced [7] and evaluated at the word level. This study examines how the model should be used in a voice agent and which other cues it should be combined with. End-of-turn detection is evaluated at 1,211 silence onsets in Real-TurnTurk [8] with recording-disjoint and speaker-disjoint folds and paired cluster bootstrap tests, comparing a fine-tuned, a zero-shot and a deployed text scorer. The study further analyses calibration and the latency–interruption trade-off of a silence-timeout policy, combines the text score with voice-activity timing and facial features through late fusion, quantifying the contribution of each source, and repeats the evaluation on automatic transcripts produced by Whisper large-v3 [9].

## 2 Related Work

Syntactic and pragmatic completion, intonation, final lengthening, intensity and gaze all indicate that a turn may be about to end, and a combination of these cues is more reliable than any one of them [5]. Speakers tend to gaze at the listener when yielding the turn and to avert their gaze when holding it [6].

Raux and Eskenazi [2] formulated the choice of when to respond as a trade-off between latency and interruptions and replaced fixed thresholds with a learned policy. Continuous models predict future voice activity from acoustic and lexical input [10, 11]; Voice Activity Projection learns such patterns from audio without supervision [12] and has been run in real time and in several languages [13, 14].

TurnGPT adds a turn-shift token to a pre-trained language model and reads the probability of that token after each word [15]. Production agents apply small text classifiers of this kind to the recogniser's transcript when silence is detected; the multilingual detector of LiveKit lists Turkish among its supported languages [4]. For Turkish, a synthetic [16] and a real multimodal corpus [8] are available, and TurnTurk-GPT [7] applies the TurnGPT approach to a monolingual Turkish GPT-2 [17]. That model was evaluated at the word level using text alone. In this study it is evaluated at silence onsets and on automatic transcripts, and combined with non-textual cues.

## 3 Data and Decision Points

### 3.1 Corpus

Real-TurnTurk comprises eleven unscripted dyadic video calls, six of which are publicly released.[1] This study uses the released subset, which contains six calls between six Turkish speakers, 1.94 h in total, with each participant on a separate audio channel. It provides per-channel voice activity from Silero VAD [18] on a 62.5 ms grid, time-aligned reference transcripts, backchannel annotation and, on the same grid, facial features of both participants from the MediaPipe Face Landmarker [19, 20]: 478 landmarks including the irises, head pose and 52 blendshape coefficients per frame. Consecutive speech blocks of one speaker form a turn, and a backchannel by the listener does not end it. This gives 578 turns and 15,522 word-level decision points.

[1]https://huggingface.co/datasets/tugrulbayrak/Real-TurnTurk

### 3.2 Silence Onsets

A silence onset is the end of a speech block. After a turn end the other participant takes the floor, and after a turn-internal pause the same speaker resumes. The 578 turn ends are used as positives and the 633 turn-internal pauses aligned to the preceding word as negatives. The pauses have a median duration of 0.50 s, and 297 of them last longer than 0.5 s. With this class ratio, a random scorer reaches an average precision (AP) of 0.477. For each onset, the time $t$ at which silence begins, the current speaker and, for internal pauses, the pause duration are stored. All features use only information available before $t$.

## 4 Systems

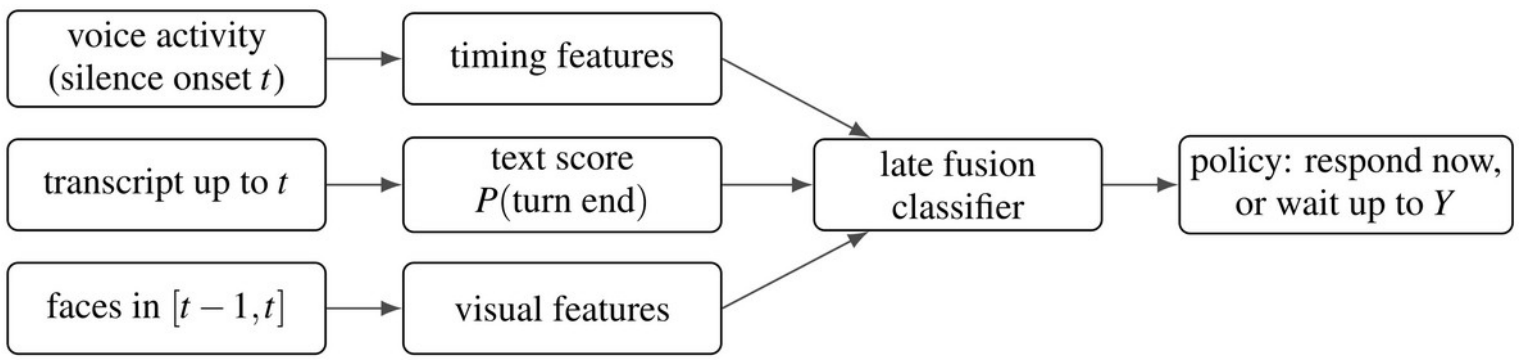

**Fig. 1.** Decision at a silence onset

### 4.1 Text Scorers

TurnTurk-GPT extends the Turkish GPT-2, a model with 124M parameters, with a turn-shift token and two speaker tokens; it is pre-trained on synthetic Turkish dialogue[2] and fine-tuned on Real-TurnTurk with the causal language modelling objective [15]. The input is lower-cased and unpunctuated, and the score of a word is the probability of the turn-shift token following it. In the recording-disjoint setting the model is fine-tuned without the test recording, and in the speaker-disjoint setting without any recording of the test speakers, so that it is evaluated on unseen users. Scores are averaged over three seeds.

The zero-shot baseline is the same GPT-2 without fine-tuning, reading the same words with turns separated by line breaks; its score is the probability of a line break after the word, the most effective of four zero-shot readings in [7]. The multilingual turn detector of LiveKit [4] is used as released, with its own normalisation, context window and inference code.

[2] https://huggingface.co/datasets/tugrulbayrak/Syn-TurnTurk

### 4.2 Timing and Visual Features

Only information about the current speaker is used, since in a deployed system the listener is the agent itself. Four timing features are computed from voice activity at $t$: the duration of the most recent speech block, the time the speaker has held the floor in the current turn, the number of speech blocks in the turn, and the time since the other participant last held the floor.

The visual features use nine MediaPipe signals of the speaker, namely horizontal gaze, head yaw and pitch, mouth and jaw opening, blink, downward gaze, smile and inner brow raise, each $z$-normalised per person over the recording. From these, the mean of each signal over $[t - 1, t]$, its change in the last 0.25 s and the variability of head motion are computed, giving 19 features in total.

### 4.3 Fusion and Evaluation

As shown in Fig. 1, feature sets are combined with the logit of a text score in a logistic regression (LR) and in a gradient-boosted tree ensemble (GB) with a depth of 3 and 200 iterations [21], trained with the same folds as the text model on the silence onsets of the training recordings only. Ranking is measured with AP and AUC, and calibration with the Brier score [22] and the expected calibration error (ECE) [23]. Confidence intervals are percentile intervals from 2,000 cluster bootstrap replicates that resample the six recordings [24]. Systems are compared on the same replicates; a difference is significant when its 95% interval excludes zero.

## 5 Results

### 5.1 Text Scores at Silence Onsets

Table 1 reports AP with its 95% confidence interval and AUC at the 1,211 silence onsets, where a random scorer reaches an AP of 0.477; the last column gives the paired AP difference to TurnTurk-GPT[3], and an asterisk marks a difference whose interval excludes zero. On held-out recordings the fine-tuned model obtains the highest AP among the text scorers. Its AP is 0.035 higher than that of the zero-shot model and 0.095 higher than that of the deployed detector, both significantly, and it is ahead of the zero-shot model in each of the six recordings. On unseen speakers the fine-tuned model retains a significant margin of 0.049 over the deployed detector and does not differ significantly from the zero-shot model, which indicates that its advantage over the multilingual system is not limited to familiar speakers. As the words before a pause often form a complete clause even when the speaker continues, the text score is combined with other cues in Section 5.4.

[3] https://huggingface.co/tugrulbayrak/TurnTurk-Gpt

**Table 1.** Text scorers at silence onsets

| System | AP [95% CI] | AUC | ΔAP |
|---|---|---|---|
| *Recording-disjoint* | | | |
| TurnTurk-GPT | .625 [.551,.667] | .638 | – |
| Zero-shot LM | .590 [.510,.635] | .630 | −.035* |
| LiveKit detector | .530 [.455,.579] | .537 | −.095* |
| *Speaker-disjoint* | | | |
| TurnTurk-GPT | .579 [.502,.633] | .594 | – |
| Zero-shot LM | .590 [.510,.635] | .630 | +.011 |
| LiveKit detector | .530 [.455,.579] | .537 | −.049* |

### 5.2 Calibration

A fixed-threshold policy requires probabilities with the same meaning across conversations. Table 2 lists calibration measures over all 15,522 words, where the observed turn-end rate is 0.037; lower values are better, and asterisks mark Brier scores significantly higher than that of TurnTurk-GPT on held-out recordings. The fine-tuned model has the lowest Brier score, whereas the mean probability of the deployed detector is approximately twice the observed turn-end rate. Trained on real data only, without synthetic pre-training, the same model assigns probabilities close to zero to nearly all words, with a mean of 0.000 and a log-loss of 0.553, and a usable threshold is difficult to set. This suggests that synthetic pre-training makes its output more interpretable as a probability.

**Table 2.** Calibration

| System | Brier | Log-loss | ECE | Mean *p* |
|---|---|---|---|---|
| TurnTurk-GPT, rec.-disj. | .0346 | .165 | .019 | .028 |
| TurnTurk-GPT, spk.-disj. | .0370 | .183 | .027 | .034 |
| Zero-shot LM | .0352* | .147 | .017 | .042 |
| LiveKit detector | .0488* | .195 | .057 | .079 |
| Real data only | .0372* | .553 | .037 | .000 |

### 5.3 Latency and Interruptions

The simulated policy follows Fig. 1. Once silence has lasted $X$ seconds, the agent queries the model and responds if the score exceeds a threshold, and otherwise after $Y$ seconds at the latest; the delay at a turn end is therefore either $X$ or $Y$. At an internal pause of duration $d$ the agent interrupts if $d > X$ and the score exceeds the threshold, or if $d > Y$. The baseline responds after a fixed silence $T$. Each curve is the lower envelope over $X \in \{0.1, \ldots, 0.6\}$ s, $Y \in [0.3, 3]$ s and the threshold.

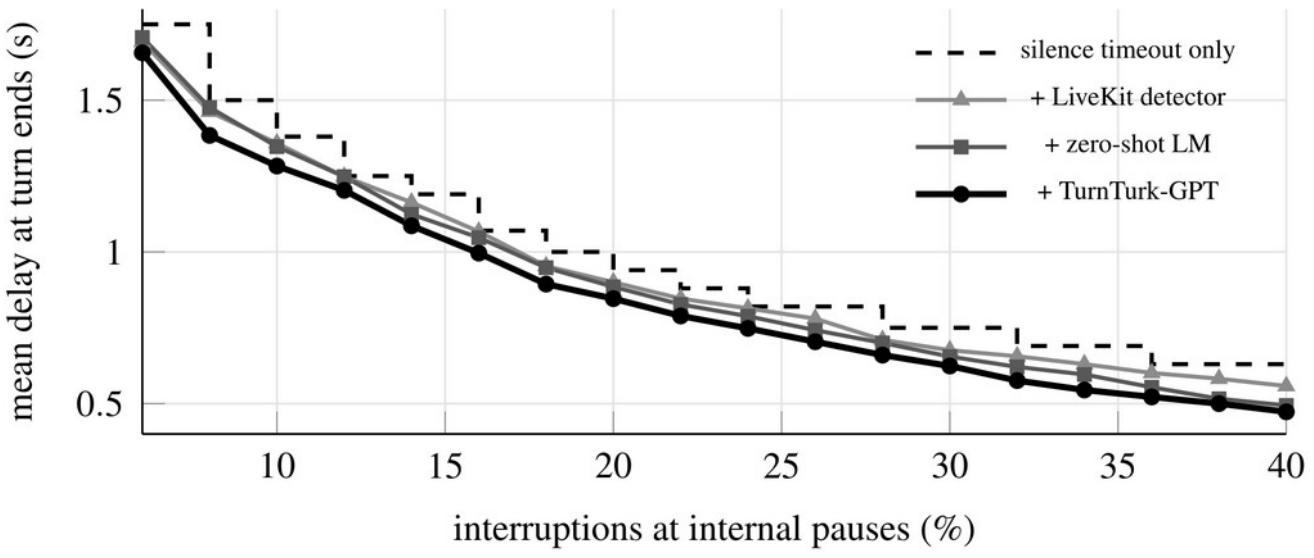

**Fig.** 2**.** Latency-interruption trade-off

Fig. 2 plots the mean delay at turn ends against the interruption rate at internal pauses for recording-disjoint scores; lower values indicate better performance. With a fixed timeout, interrupting at most 10% of internal pauses requires a delay of 1.38 s at every turn end, since the median duration of internal pauses is 0.50 s. The fine-tuned text score reduces the delay by 0.097, 0.089 and 0.094 s at the 10%, 15% and 20% budgets, all significant over 1,000 bootstrap replicates. It is also significantly shorter than with the deployed detector, by 0.074 and 0.054 s at the 15% and 20% budgets, and than with the zero-shot model, by 0.040 s at the 20% budget. With speaker-disjoint scores the fine-tuned model likewise yields the shortest delays among the text scorers at the 10–20% budgets, with delays of 1.29, 1.06 and 0.87 s.

### 5.4 Combining Text with Timing and Facial Cues

Table 3 reports AP at the silence onsets for logistic regression and gradient boosting on both splits. The four timing features of the speaker reach 0.615–0.652; as they are derived from voice activity alone, they are an inexpensive complement to the text score. On held-out recordings, adding the fine-tuned text score to timing raises the AP significantly, by 0.029 with LR and 0.030 with GB; this combination is the best timing-based system with both classifiers and, with LR, significantly exceeds timing combined with the zero-shot model by 0.026 and timing combined with the deployed detector by 0.027. On unseen speakers it reaches the highest AP with GB, 0.630.

The speaker's facial features reach 0.544–0.626 on their own. With GB on held-out recordings they raise the AP of timing significantly, by 0.044, and adding a text score gives 0.689–0.700. With LR on held-out recordings, the fine-tuned text score adds a significant 0.028 AP to timing and facial features and exceeds the zero-shot model by 0.006 and the deployed detector by 0.029, both significantly.

**Table 3.** Late fusion at silence onsets

| | Rec.-disjoint | | Spk.-disjoint | |
|---|---|---|---|---|
| Features | LR | GB | LR | GB |
| Text only: TurnTurk-GPT | .625 | .625 | .579 | .579 |
| Text only: zero-shot LM | .590 | .590 | .590 | .590 |
| Text only: LiveKit | .530 | .530 | .530 | .530 |
| Timing | .644 | .652 | .622 | .615 |
| Timing + LiveKit | .646 | .672 | .637 | .622 |
| Timing + zero-shot LM | .647 | .674 | .642 | .614 |
| Timing + TurnTurk-GPT | .673 | .682 | .631 | .630 |
| Visual | .572 | .626 | .544 | .549 |
| Timing + visual | .644 | .695 | .626 | .605 |
| Timing + visual + LiveKit | .643 | .700 | .626 | .607 |
| Timing + visual + zero-shot LM | .666 | .689 | .637 | .613 |
| Timing + visual + TurnTurk-GPT | .672 | .698 | .631 | .613 |

### 5.5 Automatic Transcripts

A deployed agent receives recogniser output rather than reference transcripts, so each speaker channel was also transcribed with Whisper large-v3 [9], using word timestamps and voice-activity filtering. The word error rate against the reference transcripts is 19.4%. The recognised words were assigned to the same voice-activity turns, and the text models were trained on the reference transcripts as before and applied to the recognised text of the held-out recording. On recognised text the procedure yields 664 turn ends, 546 aligned pauses and 15,008 words. As the decision points differ between the two conditions, the comparison across conditions is based on AUC.

**Table 4.** Reference and Whisper transcripts

| | Reference | | Whisper | |
|---|---|---|---|---|
| System | AP | AUC | AP | AUC |
| TurnTurk-GPT, rec.-disj. | .148* | .783 | .148* | .748 |
| TurnTurk-GPT, spk.-disj. | .120 | .755 | .122 | .715 |
| Zero-shot LM | .130 | .787 | .136 | .747 |
| Random | .037 | .500 | .044 | .500 |

Table 4 compares AP and AUC on all words for the two kinds of transcript; an asterisk marks a significant improvement over the zero-shot model on the same transcript. Recognition errors lower the AUC of the fine-tuned model by 0.035, while its AP remains at 0.148. Its advantage over the zero-shot model on held-out recordings is preserved, at 0.012 AP with a 95% interval from 0.004 to 0.030. At silence onsets on recognised text, where a random scorer reaches an AP of 0.549, the fine-tuned model reaches an AP of 0.748, compared with 0.736 for the zero-shot model and 0.688 for the deployed detector; the difference of 0.060 to the deployed detector is significant.

## 6 Discussion

Timing information is available without additional processing in any agent that runs voice activity detection, and it complements the text score, since on held-out recordings neither source alone reaches the AP of their combination. The text score reflects whether the words produced up to the pause form a complete utterance, whereas the timing features reflect how long the speaker has held the floor and how the current pause relates to the preceding speech blocks.

The behaviour of the text scorers is consistent with the structure of Turkish. Because Turkish is verb-final, the evidence that an utterance is complete is concentrated in the suffixes of its final word, and removing these suffixes from the turn-final word was shown to reduce the AP of TurnTurk-GPT by up to 44% [7]. The same property explains why pauses are difficult for a text-only model, because a speaker who pauses after a complete clause and then continues produces a word sequence that is grammatically finished. At such points the text score alone cannot separate the two cases, and cues that do not depend on the words, such as the duration of the current speech block, provide the additional information.

## 7 Conclusion

This study evaluated Turkish end-of-turn detection at the point where a voice agent makes its decision, the onset of silence. On held-out recordings, a fine-tuned TurnGPT-style Turkish model separates turn ends from pauses better than a zero-shot language model and a deployed multilingual detector, with an AP of 0.625 against 0.590 and 0.530. Its synthetic pre-training is also associated with the lowest Brier score, which makes its output better suited to comparison with a fixed threshold across conversations. In a latency simulation, the fine-tuned score also shortens the response delay of a silence-timeout policy by approximately 0.09 s at the same interruption rate, and with speaker-disjoint scores it likewise yields the shortest delays among the three text scorers at the 10–20% interruption budgets. The four timing features of the current speaker, which any agent running voice activity detection obtains without additional processing, are informative on their own and complement the text score, and on held-out recordings timing combined with the fine-tuned model is the best timing-based system with both classifiers. Finally, on automatic Whisper transcripts with a word error rate of 19.4%, the fine-tuned model retains its advantage over the zero-shot model. These results point to a simple configuration for Turkish voice agents, in which the silence timeout is kept as a fallback, a causal Turkish text score is updated word by word, and at the onset of silence it is combined with timing features of the speaker.

Adapting the model to a deployed agent's own users is a natural next step towards further improvement, since the gains from fine-tuning and facial features are largest for speakers represented in the training data. Future work will extend the corpus to more recordings and speakers in order to strengthen these gains for new users, and will also evaluate the policy in live interaction with a streaming speech recogniser.